\documentclass[letterpaper, 10 pt, conference]{ieeeconf}  

\IEEEoverridecommandlockouts                              
\usepackage[utf8]{inputenc}
\usepackage[T1]{fontenc}
\usepackage{graphicx}
\usepackage{float}
\usepackage{subfigure}
\usepackage{amsmath}
\usepackage{algorithm}
\usepackage{algpseudocode}
\usepackage{booktabs}
\usepackage{vcell}
\usepackage[table,xcdraw]{xcolor}

\title{\LARGE \bf
GSECnet: Ground Segmentation of Point Clouds for Edge Computing
}

\author{
    Dong He$^{1}$, 
    Jie Cheng$^{2}$, 
    and 
    Jong-Hwan Kim$^{1}$,~\IEEEmembership{Fellow,~IEEE}
\thanks{$^{1}$Dong He and Jong-Hwan Kim are with the School of Electrical Engineering, Korea Advanced Institute of Science and Technology, Daejeon, Republic of Korea 
        {\tt\small \{d.he, jhkim\}@kaist.ac.kr}}%
\thanks{$^{2}$Jie Cheng is with the Department of Computer Science, Chongqing University of Technology, Chongqing, PR China.
        {\tt\small 1071568718@2018.cqut.edu.cn}}%
\thanks{*Dong He and Jie Cheng contributed equally to this work.}%
}

\begin{document}

\maketitle
\thispagestyle{empty}
\pagestyle{empty}

\begin{abstract}
Ground segmentation of point clouds remains challenging because of the sparse and unordered data structure. This paper proposes the GSECnet - Ground Segmentation network for Edge Computing, an efficient ground segmentation framework of point clouds specifically designed to be deployable on a low-power edge computing unit. First, raw point clouds are converted into a discretization representation by pillarization. Afterward, features of points within pillars are fed into PointNet to get the corresponding pillars feature map. Then, a depthwise-separable U-Net with the attention module learns classification from the pillars feature map with an enormously diminished model parameter size. Our proposed framework is evaluated on SemanticKITTI against both point-based and discretization-based state-of-the-art learning approaches, and achieves an excellent balance between high accuracy and low computing complexity. Remarkably, our framework achieves the inference runtime of 135.2 Hz on a desktop platform. Moreover, experiments verify that it is deployable on a low-power edge computing unit powered 10 watts only.

\end{abstract}

\section{INTRODUCTION}
Recently, along with the extensive utilization of LiDAR sensors in various applications such as AR, autonomous driving, increasing impressive outputs of semantic segmentation on 3D point clouds gain wide attention from the academy and industry community. In particular, high-efficiency ground segmentation for point clouds is of considerable crucial. However, unlike 2D images with a dense and organized structure, point clouds captured from the LiDAR sensor are sparse, unevenly distributed, and unordered by nature. These specific properties impose enormous challenges to extract useful information from point clouds. In KITTI \cite{geiger2013vision} dataset, Velodyne HDL-64E LiDAR has been equipped to collect point clouds of surroundings. Millions of points are received every second, yet over 1/3 of which are reflected from the ground. On the contrary, person, rider, vehicular, and bicycles are less than 1\% in most frames according to SemanticKITTI \cite{behley2019semantickitti} dataset. It is obvious that applications dealing with point clouds suffer performance deterioration because of abundant ground points. Therefore, ground segmentation is not only used to generate terrain models but also acted as a crucial pre-process since post-processes of applications depend on the validity of the segmentation result. It also appears that in the wake of point cloud data becoming higher definition and growing demand for deploying on more edge computing units, the novel ground segmentation task needs to take into account both high accuracy and low computing complexity. 
    \begin{figure}[tp]
      \centering
      \includegraphics[width=0.47\textwidth]{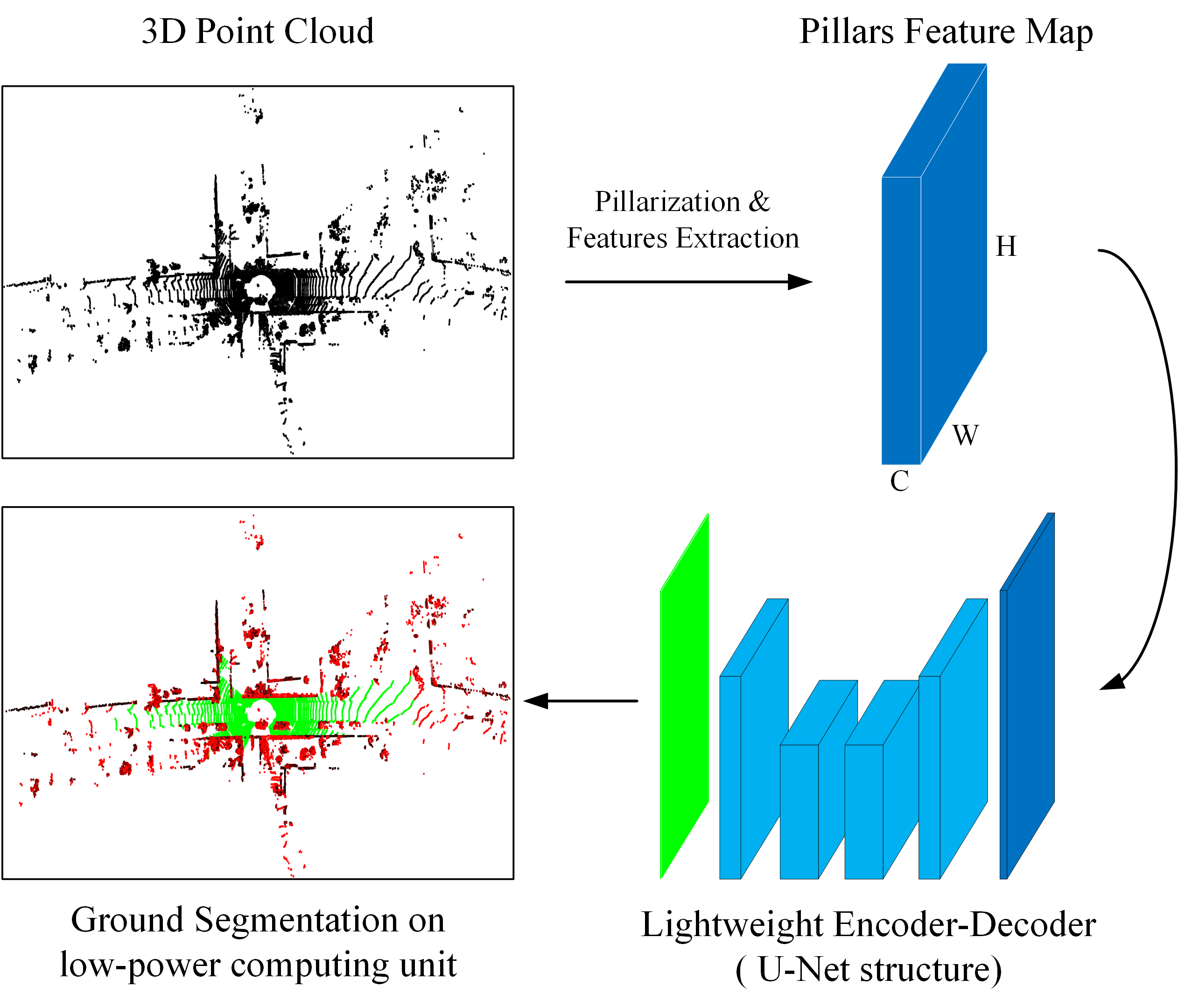}
      \caption{A schematic overview of our framework. Input is the 3D point cloud; Outputs are points with ground/non-ground labels. Points can be estimated from pillars feature map on a low-power edge computing unit through the efficiently optimized U-Net structure.  }
      \label{figurelabel}
   \end{figure}

Heuristic-based methods are first introduced, in which hand-crafted geometric features are applied to estimate ground panels then segment ground points. With the rise of deep learning research, learning-based methods are developed rapidly and offer a promising performance. However, 3D CNN operation suffers high computing complexity because of unnecessary computation in the sparse region. Therefore, some pioneer works represent point clouds into an organized form that can apply 2D CNN operation. In our work, we use pillar representations where points are discretized with a specific resolution on the x and y-axis, then apply 2D semantic segmentation.

In 2D semantic segmentation, U-Net \cite{ronneberger2015u} is one of the leading approaches, which provides several benefits: U-Net allows for the use of global location and context at the same time; it works with a few training samples and yields good performance for segmentation tasks; an end-to-end pipeline processes the entire image in the forward pass and directly produces segmentation maps. These benefits assure that U-Net preserves the input images' entire context, which is a primary improvement compared to patch-based segmentation approaches. Hence, we apply U-Net as the backbone in our work. 

In this paper, our goal is to find a method that would perform ground segmentation from raw point clouds on a low-power edge computing unit. After reviewing the methods mentioned above, we first propose the network, GSECnet as shown in Fig. 1, which uses the U-Net as the backbone and equips with depthwise-separable convolution \cite{chollet2017xception} and attention module \cite{woo2018cbam}. We successfully cut down the model parameters to 1\% of the original U-Net implementation. Another novel idea of our framework is the full pillar-wise prediction, which significantly reduces the computing complexity. Meanwhile, we study the normal of points ignored by ground segmentation works, and find that it boosts accuracy with slightly increasing computing complexity in inference. To investigate the performance of GSECnet, we train and test the model on SemancticKITTI. Compared to top methods: PointNet++ \cite{qi2017pointnet++} and GndNet \cite{paigwar2020gndnet}, we show an excellent balance between high accuracy and low computing complexity.

Our contributions are threefold:
\begin{itemize}

\item We present an end-to-end deep neural network framework called GSECnet for ground segmentation of point clouds, workable on a low-power edge computing unit.
\item We adopt the normal feature of points and distribution controlled undersampling to the framework and observe improvements in experiments.
\item We use full pillar-wise prediction to achieve lower computing complexity.

\end{itemize}

The rest of the paper is organized as follows. In Section II, we review related work; Section III describes the proposed method and implementation process; Evaluation on semanticKITTI dataset and ablation study are shown in Section IV; Finally, concluding remarks and future work follow in Section V.
 
\section{RELATED WORK}
Mainly due to the particular aspects of the 3D point cloud, many meaningful attempts of ground segmentation have been performed. These related works can be roughly categorized into heuristic methods and learning methods. This section briefly surveys heuristic methods and discusses their limitations, such as weak adaptability in practices. Then we review learning methods from two types of data representation: point-based and discretization-based \cite{guo2020deep}.

\subsection{Heuristic ground segmentation}
Traditionally, earlier works that appeared for ground estimation mainly rely on hand-crafted features from geometrical constraints and statistical rules. The ground plane method is fitted to the extracted geometric information of points using a model-fitting algorithm. The elevation map approach \cite{thrun2006stanley} projects a 3D point cloud to a 2.5D grid, then the maximum and minimum threshold values of elevation are manually assigned. This approach cannot handle multiple horizontal surfaces such as bridges, tunnels, or treetops. Gallo \cite{gallo2011cc} leverages RANSAC to fit the ground plane, which has not dealt with vertical panels of buildings. Besides, Liu \cite{liu2019ground} integrates Gaussian process regression and robust locally weighted regression to model the ground plane, resulting in high computational complexity. Wolf \cite{wolf2015fast} develops a framework to capture geometric features of segmentations by a pre-trained CRF. These methods with geometric constraints perform well in their defined scenes, but various limitations occur in realistic scenes. In short, heuristic ground segmentation approaches show shortcomings in adapting to all cases simultaneously.

\subsection{Learning-based ground segmentation}
More recently, learning-based approach has focused on attention, which uses the neural network to classify points by learning a feature representation. Two groups of representative works are reviewed below.

\textbf{Point-based methods} directly take raw point clouds and output segmentation results. As the pioneering work, PointNet \cite{qi2017pointnet} presents a deep learning architecture that can directly handle point sets. It models each point with shared MLPs and then aggregates a global feature using a symmetric aggregation function, which achieves outstanding segmentation results. PointNet++ \cite{qi2017pointnet++} builds a hierarchical structure to extract the local contextual features between points. Some works contributed to constructing an ordered feature sequence for convolution operation. PointwiseCNN \cite{hua2018pointwise} focuses on defining the point convolution operations. PyramidPoint \cite{varney2020pyramid} employs a dense pyramid structure instead of a U-Net. However, these point-based methods need the high computational cost neighbor searching algorithm to obtain neighboring information \cite{guo2020deep}, which is problematic to be applied on edge computing units.
\begin{figure*}[ht]
      \centering
      \includegraphics[width=0.95\textwidth]{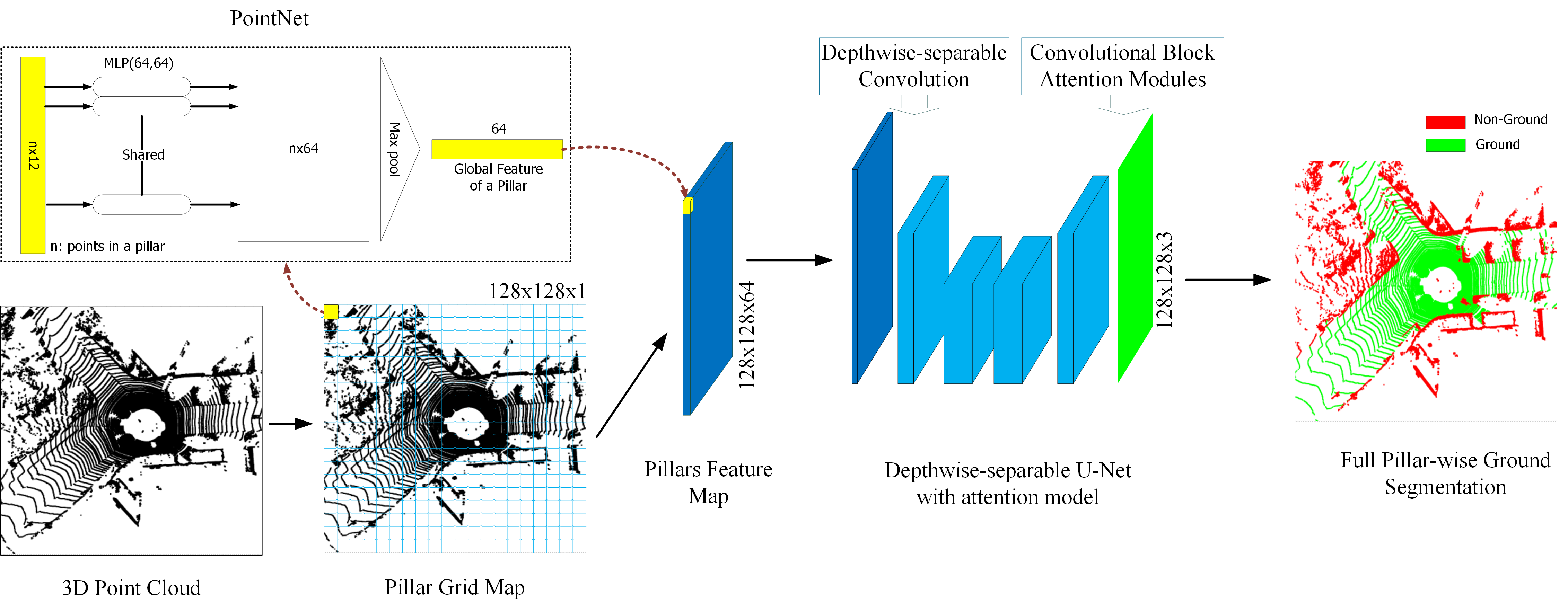}
      \caption{Overall structure. 1) Representing a cloud point by pillarization: 3D point cloud is discretized into a 128x128x1 pillar grid map. 2) Generating Pillars Feature Map: Points with 12 features (original features of points, normals of points, and pillars location information) in each pillar are fed into PointNet to generate Pillars Feature Map. 3) Estimating ground points: Pillars Feature Map goes through the encoder-decoder to estimate ground pillars with low computing complexity and high accuracy.}
      \label{figurelabel}
\end{figure*}

\textbf{Discretization-based methods} emerge as the alternative with high efficiency, which convert a point cloud into the ordered discrete structure such as lattice, voxel, or pillar. SPLATNet \cite{su2018splatnet} interpolates point clouds to a permutohedral sparse lattice then executes 3D CNN. VoxelNet \cite{zhou2018voxelnet} discretizes point clouds into voxels and uses dense 3D convolutions. Nevertheless, it is difficult for these 3D CNN frameworks to gain both accuracy and efficiency. To reduce the computational cost of 3D CNN, PointPillars \cite{lang2019pointpillars} and pillar-based detection \cite{wang2020pillar} use pillars instead of voxels and encode features of the point cloud to the pseudo image, and then apply CNN 2D. Paigwar \cite{paigwar2020gndnet} adopts an analogous way but uses a CRF-based elevation map as ground truth to learn each pillar's height, which can estimate and model ground at a speed of 55 Hz on KITTI dataset. Inspired by these discretization-based methods, in our preliminary work, PointNet was employed to generate the pillars feature map from point clouds.

\section{PROPOSED FRAMEWORK: GSECnet}
In this section, we first introduce the overall structure of GSECnet, as depicted in Fig. 2. We delineate the generation of pillars feature map by passing through pillarization and PointNet, the design of depthwise-separable U-Net with attention module, and the process of data undersampling from raw point clouds.

\subsection{Point cloud pillarization} 

Because of point clouds' sparsity aspect, all points must be inspected individually to determine whether some of them belong to the interesting area; meanwhile, copious void pillars exist after pillarization. In this work, we plan an appropriate pillar size to guarantee both accuracy and efficiency. In the PointPillars authors' point of view, the \(z\) dimension is needless because it largely increases computational complexity while not contributing to the segmentation result. Following the pipeline of pillar-based approaches, we discretize the environment into a pillar grid map (128, 128, 1), and the size of each pillar is 0.8m x 0.8m x 8m. Entire environment ranges of \(x\), \(y\), and \(z\) dimensions are [[-51.2, 51.2], [-51.2, 51.2], [-4, 4]] meters, and outliners are removed. Meanwhile, we specify the maximum number of points per pillar as 64 by applying random undersampling.

\subsection{Data augmentation}
   
Similar to PointPillars and GndNet, all points in each pillar are augmented. In particular, normal features of points are appended. In Fig. 3, we observe that normals of most ground points are organized and orthogonal to the horizontal plane. On the contrary, objects points' normals randomly orientated to diverse directions. Clearly, appended normal features will assist our classification network to segment ground points correctly. 

Given a point \( p_i(x_i, y_i, z_i) \), we select \(k\) nearest points, denoted as \( {Q_i \{q_{i1} ,q_{i2},...,q_{ik}\} }\) by using KD-tree \cite{bentley1975multidimensional}, and then apply least squares method to fit the part of plane with selected points. Thereafter, normals of selected points can be obtained. The process is given by 

\begin{equation}
a x+b y+c z+d=0,
\end{equation}

\begin{equation}
X=\left[\begin{array}{ccc}
x_{q_{i 1}} & y_{q_{i 1}} & 1 \\
x_{q_{i 2}} & y_{q_{i 2}} & 1 \\
\vdots & \vdots & \vdots \\
x_{q_{i k}} & y_{q_{i k}} & 1
\end{array}\right] \quad, \quad Y=\left[\begin{array}{c}
z_{q_{i 1}} \\
z_{q_{i 2}} \\
\vdots \\
z_{q_{i k}}
\end{array}\right],
\end{equation}

\begin{equation}
\left[\begin{array}{r}
-\frac{a}{c} \\
-\frac{b}{c} \\
-\frac{d}{c}
\end{array}\right]=\left(X^{T} X\right)^{-1} X^{T} Y,
\end{equation}

\begin{equation}
\begin{aligned}
& V_{i}=\left[-\frac{a}{c},-\frac{b}{c}, 1\right], and \\
& N_{i}=\frac{V_{i}}{\left|V_{i}\right|}.
\end{aligned}
\end{equation}
from (1), equation of line for space, we get (2), \(k\) points equation in space. In (3), the normals of \(p_i\) are calculated, while normalization is conducted in (4). \(V_i\) is the normals before normalization; \(N_i\) is the normals of a point \(p_i\). This process is similarly implemented in open3D \cite{zhou2018open3d}. Eventually, the original 4-dimensional of point are augmented to 12-dimension \( \{ x , y, z, i, x_c, y_c, z_c, x_p, y_p, x_n, y_n, z_n\} \), where \(x,y,z\) denote the coordinates; \(i\) the intensity; \(x_c,y_c,z_c\) distance to the mean of all points in the pillar; \(x_p,y_p\) the offset from the pillar center; and \(x_n,y_n,z_n\) the normals.
\begin{figure}[hpt]
      \centering
      \includegraphics[width=0.42\textwidth, height=7cm]{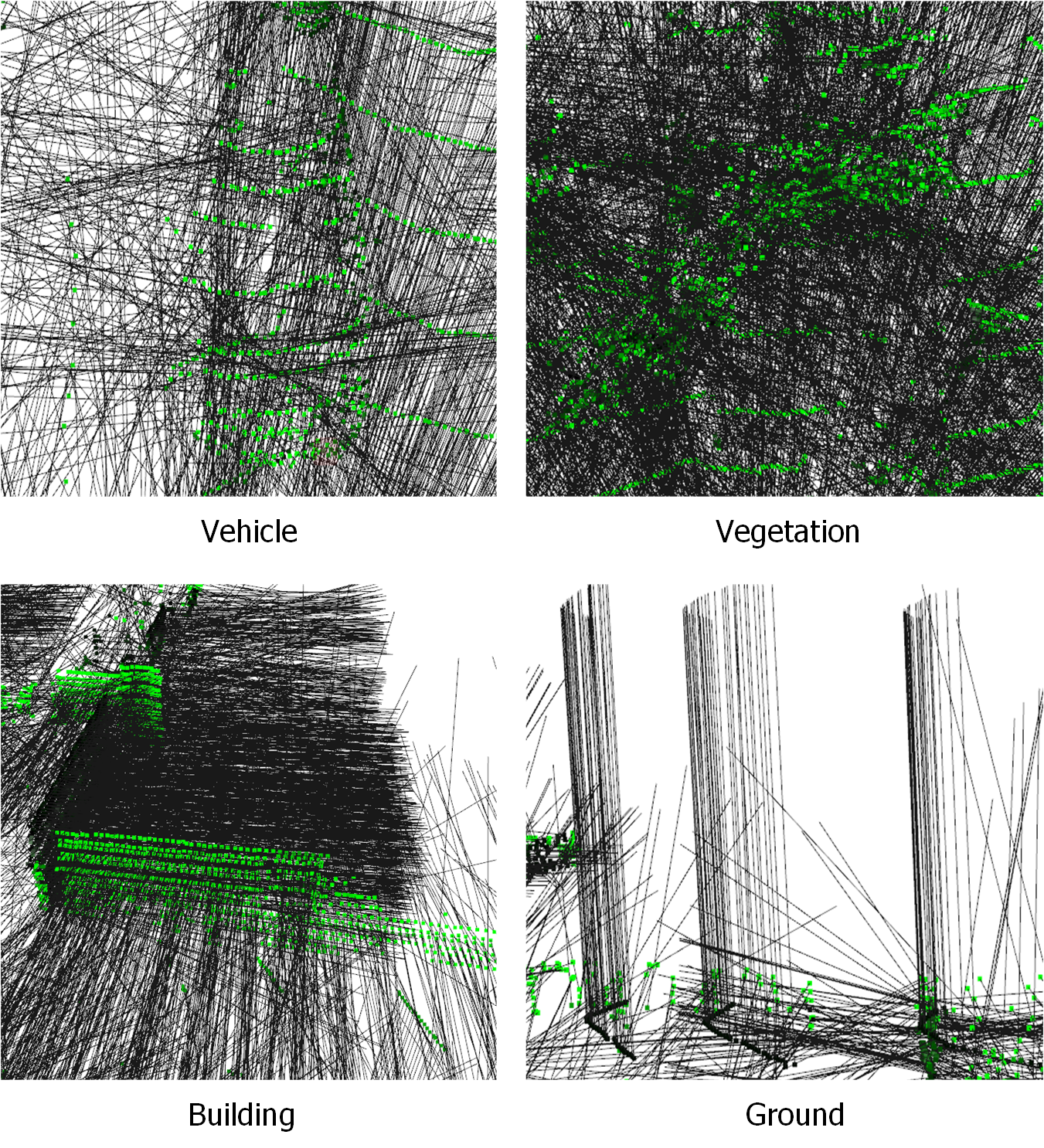}
      \caption{Visualization for normals of 4 major categories. Most ground points are orthogonal to horizon terrain; other top-3 category points orientate randomly.}
      \label{figurelabel}
   \end{figure}
   
\subsection{Pillars feature maps}
Taking advantage of the simplicity and powerful representation capability of PointNet, we apply it to aggregate features of pillars. A simplified PointNet is adopted in our work, which comprises a linear layer with the batch norm and ReLu. Augmented point features are leveraged to generate 64 channels pseudo image of pillars, called pillars feature map, using the simplified PointNet with a max-pooling operation. Thus, the size of the pillars feature map is (128, 128, 64). This process follows the line of pillar-based approach, yet we modify the size of pillars feature map in our work.

\subsection{Depthwise-separable U-Net with attention module}
Semantic segmentation works often adopt standard U-Net. Nevertheless, the model parameter is large and not feasible to deploy on our target platform because Jetson Nano has limited memory and low computing ability. Some works such as GndNet modifies U-Net by cutting layers, yet the computing ability is still insufficient to be deployed on Jetson Nano in our experiments. Additionally, because of vacant pillars, imbalanced data lie in the pillars feature map. Thus, we supersede standard U-Net and design a depthwise-separable U-Net with attention module. This model has significantly fewer parameters, only 1\% of standard U-Net implementation, but performance is retained similar to standard U-Net in our work. A smaller with similar performance to large models is crucial for autonomous driving vehicles driven by restricted isolated power. Furthermore, it is the first time to segment ground on a Jetson Nano with the KITTI point cloud input to the best of our knowledge.

Our encoder-decoder network extends U-Net as backbone with depthwise-separable convolutions (DSCs) \cite{chollet2017xception} and convolutional block attention modules (CBAMs) \cite{woo2018cbam}. We transfer convolution operations in U-Net to DSCs to decrease the model parameters, while use CBAMs after convolution operations to enhance classification ability by avoiding empty pillars. Three encoder-decoder modules are employed in our network, as shown in Fig. 4. On an encoder side, DSC operation (yellow arrows) abstracts features with a small number of parameters. Then, features pass through CBAMs (blue arrows) to learn the inherent relationship of points and wait to concentrate via the skip-connections (grey arrows), which permits the model to use multiple scales of the input to generate the output. Meanwhile, maxpooling (cyan arrows) reduces the feature map by half. On the decoder side, a bilinear upsampling (red arrows) doubles the feature map size, and then the culminating feature maps are concatenated with the previous encoder’s output via the skip-connections. Then, DSC operation with a double convolution reduces the number of feature maps by a quarter or half. Finally, instead of a fully connected layer, the last layer is a 1×1 convolution (green arrow), which yields a single feature map showing the predicted segmentation results. Recently, the authors in \cite{trebing2021smaat} suggested a similar one and demonstrated that their model achieves comparable performance as standard U-Net for weather prediction. However, our model is around 1/10 size of their model and optimized to the pillars feature map.
\begin{figure}[t]
      \centering
      \includegraphics[width=0.42\textwidth, height=7cm]{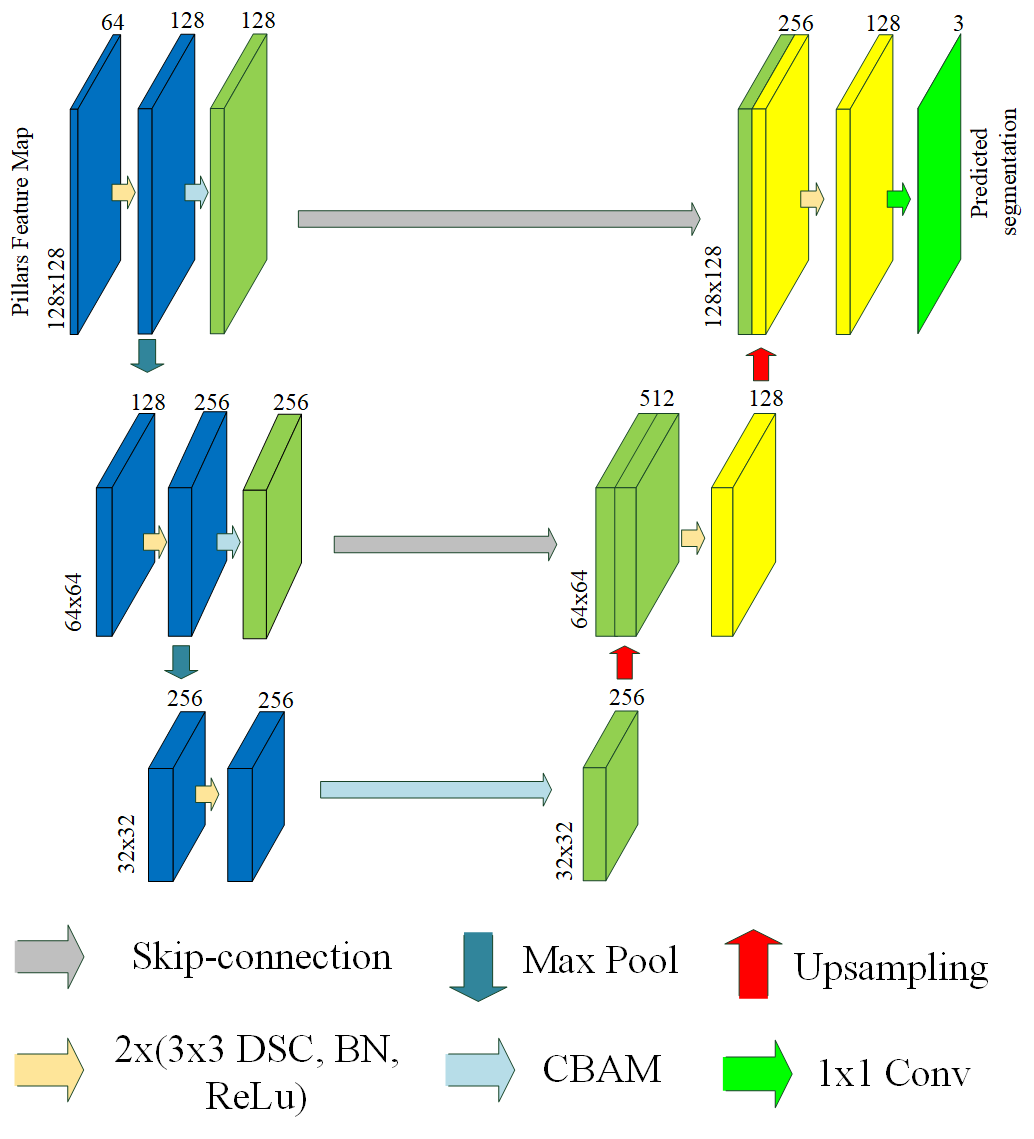}
      \caption{The proposed depthwise-separable U-Net as our encoder-decoder. Numbers of left side are the size of feature maps. Above numbers are the channels of feature maps.}
      \label{figurelabel}
   \end{figure}

\subsection{Distribution controlled undersampling}
Before pillarization, undersampling is requisite to regulate the number of input points fewer than a fair amount (e.g., 100K in our study). Since the point cloud distribution is imbalanced, points in the far region are much fewer than the adjacent region. Yet, we note that works often adopt average random undersampling, which leads to the loss of critical points in the far region. To maintain uniform distribution, we design an undersampling strategy called distribution controlled undersampling. 

For more details, we generate sections as shown in Fig. 5. The probabilities of section undersampling are determined according to both the distance from LiDAR sensor and the density of points in sections. This process is summarized in Algorithm 1. Lastly, points \(P\) are assigned the different sampling ratios according to \(S\). Although the distribution controlled undersampling is somewhat slower than average random undersampling, it shows improved ground segmentation performance stated in the ablation study.
\begin{figure}[htp]
      \centering
      \includegraphics[width=0.21\textwidth]{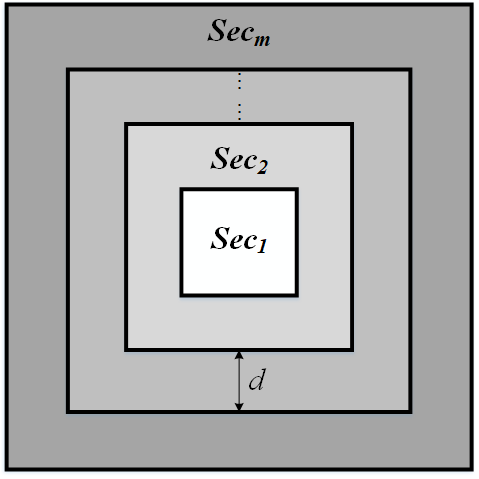}
      \caption{ Undersampling sections. The shape of sections follows the pillars features map as squared; \(d\) is manually assigned to 0.8 meters. }
      \label{figurelabel}
   \end{figure}
\renewcommand{\algorithmicrequire}{\textbf{Input:}}
\renewcommand{\algorithmicensure}{\textbf{Output:}}
\begin{algorithm}[htb]
  \caption{ Distribution controlled undersampling}
  \label{alg:Framwork}
  \begin{algorithmic}[1]
    \Require
      The set of all points in a frame (\(n\) points) \(  {P =\{p_1(x_1,y_1,z_1), p_2(x_2,y_2,z_2) ,..., p_n(x_n,y_n,z_n) \} }\); The section interval \(d\);
    \Ensure
      The set of undersampling by sections' probabilities (\(m\) sections), \(  { S=\{s_1,s_2,...,s_m \} }\); The dictionary of points to corresponding section \( \{P \rightarrow Sec\}\)
    
    \For{each $i\in [1,n]$}
      \State \( l_i = max(abs(x_i),abs(y_i)) \);
      \State \( j = l_i/d \);
      \State \( Section_j = +1\);
    \EndFor
     \State get a section list \( Sec \{Sec_1,Sec_2,...,Sec_m \} \) and dictionary \( \{P \rightarrow Sec\}\);
    \For{each $j\in [1,m]$}
      \State \( s_j = max(Sec)*2-Sec_j \);
    \EndFor
    \State get a set of sections' probabilities 
    \label{code:fram:select} \\
    \Return \( {S=\{s_1,s_2,...,s_m \} }\); \( \{P \rightarrow Sec\}\)
  \end{algorithmic}
\end{algorithm}

\subsection{Loss function}
We remark that a great number of vacant pillars remain in the pillars feature map. These easy negative samples can overwhelm training and lead to degenerate models. Hence, we utilize focal loss \cite{lin2017focal} for pillar classification and give small weights of easy negative samples to prevent wrong learning direction. The focal loss is expressed as
\begin{equation}
FocalLoss(p_t)=-a_t(1-p_t)^blog(p_t) 
\end{equation}
where \(a\) denotes a weighting factor; \(p_t\) the predicted probability for the class with ground points label; \(b\) a tunable focusing parameter. In our study, \(a\) and \(b\) are assigned 0.25 and 2, respectively. 

\section{EXPERIMENTS}
Our experiments involve two parts. First, we evaluated our model on desktop and Jetson Nano against PointNet++ and GndNet for ground segmentation on the SemanticKITTI. Next, the impact of our proposed U-Net model and normal features were investigated in the ablation study. 
\subsection{Dataset}
KITTI dataset is one of the most popular public segmentation datasets. On the base of that, SemanticKITTI contains 11 sequences of the KITTI dataset with 28 classes labeled data. Our training set comprises sequences 01, 02, 03, 04, 05, 06, 07, 09, and 10; meanwhile sequence 08 is used as the test set. Moreover, we generated the ground truth of ground pillars from classes of road, sidewalk, parking, and other ground.

\subsection{Evaluation metrics}
Ground segmentation performance was evaluated by accuracy, mIoU, and F1-score from the confusion matrix. Accuracy measures the fraction of the total points that the model accurately classifies; mIoU measures the similarity between predicted ground points and ground truth; F1-score indicates our model's performance regarding precision and recall. Metrics are defined as follows:
\begin{equation}
\begin{aligned}
\text { Accuracy } &=\frac{T P+T N}{T P+T N+F P+F N}  \\
m I o U &=\frac{T P}{T P+F P+F N} \\
\text {F1-score} &=\frac{2 T P}{2 T P+F P+F N}
\end{aligned}
\end{equation}
where TP, TN, FP, and FN correspond to the set of True Positive, True Negative, False Positive and False Negative points matches for ground points, respectively.

\subsection{Implementation details}
We trained all models on GTX2080Ti GPU and i7 6700k CPU. Raw cloud points were augmented with a batch size of 16. Adam optimizer was adopted with weight decay 0.0005. The initial learning rate was 0.003 and declined by a factor of 0.35 when the loss had stopped dropping. Loss was convergent roughly around 20 epochs, and it took 6-8 hours on our desktop to train the model. More implementation details can be found in \textit{https://sammica.github.io/gsec}

\subsection{Quantitative and qualitative evaluation}
We evaluated our model against the-state-of-art works: PointNet++ and GndNet on SemanticKITTI sequence No. 8 with the same settings. Table I depicts the results regarding the accuracy, mIoU, F1-score, and runtime. In the table, PointNet++ gains the highest accuracy, mIoU, F1-score, but runtime is the slowest. Point-based method gives low efficiency due to high complexity computing, as we explained before. GndNet shows a good balance of accuracy and efficiency, while ours outperform GndNet by a wide margin with an overall runtime of 135.2 Hz.
\begin{table} [h]
\centering
\caption{Results on SemanticKITTI sequence No.08}
{
\begin{tabular}{ccccc} \toprule
Method          & Accuracy       & mIoU           & F1-score       & Runtime                      \\ \hline
PointNet++ [6]           & \textbf{0.946}          & \textbf{0.922 }        & \textbf{0.913}          &  1.2 Hz                 \\ \hline
GndNet [7] & 0.741 & 0.698 & 0.513 &64.9 Hz                \\ \hline
GSECnet (Ours)  & 0.871          & 0.817          & 0.798          & \textbf{135.2 Hz } \\ \hline
\bottomrule
\end{tabular}}
\end{table}

To determine whether our model is deployable on a low-power edge computing unit, not on the professional deep learning platform with abundant resources. We then tested our proposed model on Jetson Nano, which seems unable to perform the KITTI point clouds semantic segmentation task. Experiments proved our model is workable and running by 0.1 Hz on Jetson Nano, while other models are not workable.

\subsection{Ablation study}
To study the performance trade-off of the depthwise-separable U-Net with attention module, we evaluated it against standard U-Net, Attention U-Net, and SegNet in GndNet, reported in Table II. Attention U-Net achieves a little higher accuracy, mIoU, and F1-score than ours. However, our implementation only has 0.27M parameters and needs 1.47GMac to infer. Smaller model size and low computing complexity permit ours to be deployable on Jetson Nano.

\begin{table} [h]
\centering
\caption{Evaluation on different U-Nets}
\resizebox{0.46\textwidth}{10mm}{
\begin{tabular}{cccccc} \toprule
Method          & Accuracy       & mIoU           & F1-score       & FLOPs             & Params         \\ \hline
Standard U-Net           & 0.860          & 0.798          & 0.790          & 16.96GMac         & 34.56M         \\ \hline
Attention U-Net & \textbf{0.875} & \textbf{0.822} & \textbf{0.802} & 17.24GMac         & 34.91M         \\ \hline
SegNet (in GndNet)  & 0.802          & 0.767          & 0.743          & 14.55GMac & 2.22M \\ \hline
Our U-Net  & 0.871          & 0.817          & 0.798          & \textbf{1.47GMac} & \textbf{0.27M} \\ \hline
\bottomrule
\end{tabular}}
\end{table}

We then tested our model with and without normal features of points on the same dataset. Table III indicates the performance improvement with normal features appended.
\begin{table} [h]
\centering
\caption{Results of without and with normal features}
{
\begin{tabular}{cccc} \toprule
Method          & Accuracy       & mIoU           & F1-score                \\ \hline
GSECnet (w/o normals) & 0.865 & 0.810 & 0.792              \\ \hline
GSECnet (w/ normals) & \textbf{0.871} & \textbf{0.817} & \textbf{0.798}             \\ \hline
\bottomrule
\end{tabular}}
\end{table}

\section{CONCLUSION}
This paper proposed GSECnet, designed to perform point clouds ground segmentation for edge computing. To this end, first, a lightweight and attentive U-Net was designed. Next, we proved that dataset augmentation with normal features permits the better performance of the classification network. Building upon these improvements, we demonstrated that our model performs an excellent trade-off between accuracy and efficiency. Furthermore, experiments confirmed that GSECnet is the only model deployable on Jetson Nano. However, the experiment on Jetson Nano showed that the runtime of GSECnet is around 0.1 Hz, which is far from being used in practice. In future work, we will solve this limitation by using a different representation of point clouds.

\section*{Acknowledgment}
This work was supported by Institute for Information \& communications Technology Promotion (IITP) grant funded by the Korea government (MSIT) (No.2020-0-00440, Development of artificial intelligence technology that continuously improves itself as the situation changes in the real world).

\addtolength{\textheight}{-12cm}   












\end{document}